\documentclass[runningheads,a4paper]{llncs}

\usepackage{amssymb}
\usepackage{graphicx}
\usepackage{algorithm}
\usepackage{amsmath}
\usepackage{algorithm}
\usepackage[noend]{algpseudocode}
\usepackage{array,multirow}
\usepackage{algpseudocode}
\usepackage{pifont}
\usepackage{graphicx}
\usepackage{url}
\usepackage{hyperref}

\authorrunning{Akhtar et al.}
\titlerunning{Unsupervised Morphological Expansion}

\begin{document}

\mainmatter  

\title{Unsupervised Morphological Expansion of Small Datasets for Improving Word Embeddings}

\author{\textbf{ Syed Sarfaraz Akhtar*\qquad  Arihant Gupta*\qquad Avijit Vajpayee* \qquad \\ Arjit Srivastava\qquad Manish Shrivastava}\\  \{syed.akhtar, arihant.gupta, arjit.srivastava\}@research.iiit.ac.in,\\ avijit@inshorts.com,\\ manish.shrivastava@iiit.ac.in}

\institute{Language Technologies Research Center(LTRC) \\
Kohli Center On Intelligent Systems (KCIS) \\
International Institute of Information Technology Hyderabad (IIIT-H) - 500032}


%
%
%


%
%

\toctitle{Lecture Notes in Computer Science}
\tocauthor{Authors' Instructions}
\maketitle

\begin{abstract}
We present a language independent, unsupervised method for building word embeddings using morphological expansion of text. Our model handles the problem of data sparsity and yields improved word embeddings by relying on training word embeddings on artificially generated sentences. We evaluate our method using small sized training sets on eleven test sets for the word similarity task across seven languages. Further, for English, we evaluated the impacts of our approach using a large training set on three standard test sets. Our method improved results across all languages.
\end{abstract}

\section{Introduction}

\let\thefootnote\relax\footnotetext{* These authors contributed equally to this work.}

Word representations are being widely used to solve problems of various areas of natural language processing. These include but are not limited to dependency parsing ~\cite{BSL:14}, named entity recognition ~\cite{MIL:04} and parsing ~\cite{SOCH:13}.

Word representations have been shown to contain syntactic as well as semantic regularities ~\cite{MSG2:13b}. Such regularities also extend to morphological relations (vector of ``ran" is close to the vector resulting from the expression ``walked - walk + run").

The basis of our approach lies in morphological treatment of text before training word embeddings. In this paper, we present a method for learning word representations which, for training, take morphological regularities into consideration by generating artificial sentences. These sentences contain automatically learned morphological variants of words in the corpus. This harnesses morphology to reduce data sparsity by improving the training of low frequency words with the help of their more common morphological variants (as word embeddings train better on high frequency words).

We show that our method performs well on small datasets of seven languages with significant increase across all languages except Arabic (discussed in section~\ref{discussion}). We further analyze the impact of our approach on a large training set of English. Our evaluations show that when being applied on large quantities of training data, this method is comparable to models trained on even much larger datasets.

The main contributions of this paper are:
\begin{itemize}
\item We achieved results comparable to models trained on much larger data sets.
\item We are releasing Hin-WS235, a hindi word similarity dataset containing 235 word pairs.
\end{itemize}

\section{Related Work}
Mikolov et. al. \cite{MSG:13a} introduced the Skip-gram (SG) and the continuous bag-of-words (CBOW) models trained on untagged text. Both of them followed a single layered feedforward architecture. CBOW's training algorithm relied on predicting the current word based on context and SG tried to predict the context using the current word. Mikolov et. al. \cite{MSG2:13b} also showed the semantic and syntactic regularities found in these embeddings.

Pennington et. al. \cite{JRC:14} introduced the GloVe (Global Vectors for Word Representations) model in which they analyzed the properties responsible for the morphological regularities in the earlier models. They combined the effects of local context window and  global matrix factorization methods resulting in a global log-bilinear regression model. The proposed model out-performed SG and CBOW on standard test sets.

Word Representations, as learned by the above methods, deal with word as the basic unit and do not exploit the morphological relations present between words. However, these relations are present in the embeddings as regularities in the vector space ~\cite{MSG2:13b}. These are specially useful for words which are unseen or have a low frequency in the training set and are not trained well. Luong et. al. \cite{MAC:13} and Botha and Blunsom \cite{BB:14} used Morfessor~\cite{CL:07} for word segmentation and used a combination of morpheme and word level models. Both of these approaches handled rare and unseen words using their basic morpheme units.  Luong et. al. \cite{MAC:13} used morphological recursive neural networks (RNNs) for constructing representation for a word using its morphemes. Botha and Blunsom \cite{BB:14} used log-bilinear models for building and combining representations of morphemes for constructing representations of rare or unseen words. Luong et. al. \cite{MAC:13} also introduced the Stanford Rare-Word Dataset which contains a large number of rare and morphologically complex words. 

In contrast to Luong et. al. \cite{MAC:13} and Botha and Blunsom \cite{BB:14} where an external morphological analyzer was used, Soricut and Och \cite{SO:15} induced morphological transformations in an unsupervised manner using SG (Skip-gram) word embeddings. These morphological transformations were represented as word pairs in the same embeddings space and they used simple vector arithmetic to calculate vectors for rare and unseen words. For example, for building the embedding for ``nationalism'', they evaluated the expression ``functionalism - function + nation''.  
\section{Datasets}

For all the models trained in this paper, we have used the Skip-gram~\cite{MSG:13a} algorithm. The dimensionality has been fixed at 500 with a minimum count of 5 along with negative sampling. These parameters are identical to the ones used by Soricut and Och \cite{SO:15} for the purpose of extracting morphological transformations (Though, they used their own implementation of Word2Vec). Data is pre-processed to replace digits and special characters to avoid sparsity.

As training set for English, we use the Wikipedia data~\cite{SW:10}. Soricut and Och \cite{SO:15} and Luong et. al. \cite{MAC:13} had used the same training corpus for their models. The cleaned corpus contains about 1 billion tokens. For German and French, we use News Crawl (Articles from 2010) released as a part of ACL 2014 Ninth Workshop on Statistical Machine Translation. For Arabic, Persian and Spanish, we used Wikipedia Monolingual Corpora(2014) which is licensed under "Creative Commons Attribution-ShareAlike 4.0 International Public License". For Hindi, we used HindMonocorp0.5~\cite{HIN:14}.

We use standard word-similarity datasets for testing. For English, we use Stanford English Rare-Word (RW) dataset~\cite{MAC:13}, the WS353 ~\cite{FT:02} and the RG65 dataset ~\cite{RG:65}. The Stanford Rare-Word dataset contains comparatively more rare words and morphological complexity than other datasets and is central to our experiments. For German, we use the Gur350 and ZG222 datasets ~\cite{ZG:06} and the German RG65 dataset. For French we use the French RG65 ~\cite{JI:11} dataset. For Spanish and Persian, we use Spanish-RG65 and Persian-RG65 test data-sets ~\cite{SPAPER:15}. 

For Hindi(having second highest number of native speakers in the world after Mandarin) we are releasing a similarity (Hin-WS235) dataset containing 235 word pairs. This data-set was created by manually translating and re-annotating the English WS353 ~\cite{FT:02} dataset. This dataset is crucial since its used for direct evaluation of word embeddings. This dataset will be helpful for future work on Hindi. Adhering to blind review policy, the link for the dataset will be provided in the final version of this paper. 

For rest of the paper, we have calculated the Spearman $\rho$ (multiplied by 100) between human assigned similarity and cosine similarity of our word embeddings for the word-pairs. 

\section{Morphological Expansion}

Since we are generating new sentences using morphology, morphological sets of words that should be replaced have to be constructed. A morphological set of a word contains its "first cousins". First cousins are pairs of words which are morphological forms of each other, are semantically similar and also one can be reached from the other by a transformation involving addition/removal/replacement either suffix or prefix. For example, "nation" and "national" are first cousins (addition), "nationalism" and "national" are first cousins (removal), "nations" and "national" are first cousins (replacement), but "nation" and "nationalism" are second cousins as they involve two transformation operations. We go about doing this by extracting morphological transformations in a manner similar to that of Soricut and Och \cite{SO:15} using Skip-Gram word embeddings trained on the text. 

All the thresholds mentioned have been decided after empirical fine tuning. Even though our experiments were computationally optimized, time and space complexities also played a part in deciding our thresholds.

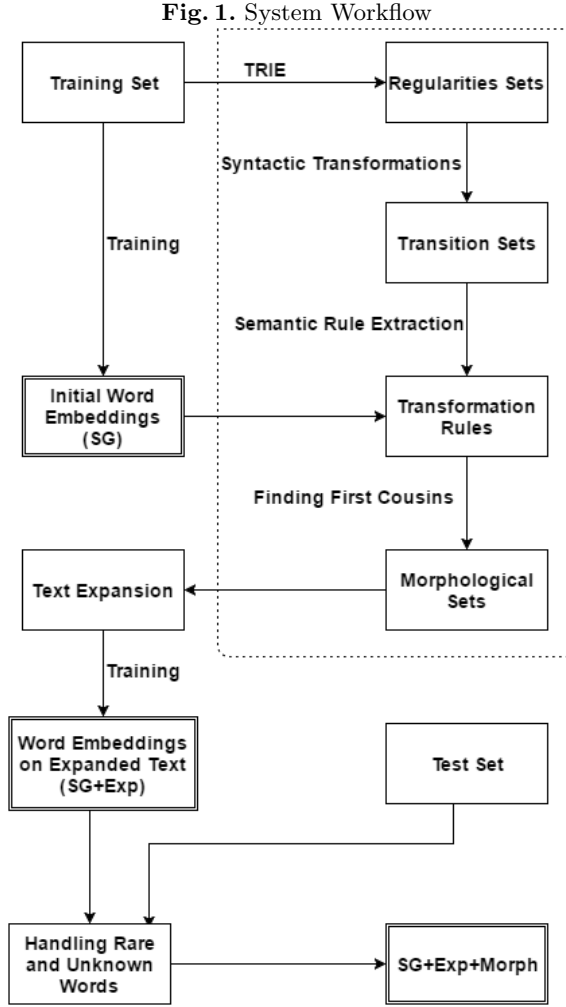
\begin{figure}[h!]
  
  \centering 
\caption{System Workflow}
\includegraphics[width = 75mm,scale=0.5]{diagram.png}
\label{workflow}
\end{figure}
In order to learn initial representations of the words, we train word embeddings (word2vec) using the parameters described above on the training set. This model is referred to as SG (Skip-gram).

\subsection{Morphological Transformations}

Morphological transformations are word pairs representing morphological regularities present at large in the corpus. Following Soricut and Och's \cite{SO:15} methodology, we extract representative candidate word pairs which exhibit same morphological behavior as at least ten other word pairs in the corpus. These candidates allow us to learn rules of morphological transformations which may not be linguistically perfect but capture orthographic regularities. We refer to the final morphological transformations extracted as ``transformation rules".

\begin{table}[h]
\begin{center}
\caption{\label{font-table8}Examples of Regularities Sets }
\begin{tabular}{|c|l|c|}
\hline \bf Regularity & \bf Set of Stems &\bf Regularity Type \\ \hline
 ed  &  reduc, unannounc, walk, ... & Suffix  \\
 ly & on, unstab, respectab, ...  & Suffix \\
 un & clear, dress, derstand, ... & Prefix \\
 dis & assemble, close, connect, ... & Prefix\\
 \hline
\end{tabular}
\end{center}
\end{table}
For extracting these transformation rules, we follow these steps:
\begin{itemize}
\item Extract {\bf Regularities} - Regularities sets (see Table~\ref{font-table8}) are the set of stems associated with each candidate prefix/suffix.

{\bf Regularities Sets}:
We use a TRIE for the extraction of candidate suffixes/prefixes. We insert all the words of the training set into the TRIE and then extract candidate prefixes. A candidate prefix is one which has more than 10 children in the TRIE. For candidate suffixes, we follow the same procedure with the only change being that the TRIE is constructed with all the words reversed. The output of this step is a candidate prefix/suffix and a set of all its stems. We call this a regularities set. As can be surmised, a number of these sets may not be linguistically correct. Also, some of these may not be true regularities and might just be data artifacts.

\item Construct {\bf Transition Sets} using these morpheme sets - Transitions sets (see Table~\ref{font-table12}) are the set of word pairs which follow the same syntactic transformation.

{\bf Transition Sets}:
There are two types of transition sets. One of them involving a prefix/suffix going to null and the other is a transition in which we have to both add and delete characters to get one word from the other. We call them null transitions and cross transitions, respectively. For evaluating null transitions, we find the intersection of each regularities set with the vocabulary of the training set. For extracting cross transitions, we find the set intersection between regularities sets. Since this leads to a large number of combinations, we evaluate only those cross sets in which both the regularities sets are large (a frequency greater than or equal to 500 for small training sets and 30000 for large training sets were used for our experiments). For both the null and cross transitions, we evaluate only those sets whose sizes are greater than 10. Also, we down-sample the transition sets thus generated to 1000 for time optimization.
\begin{table}[h]
\begin{center}
\caption{\label{font-table12}Examples of Transition Sets }
\begin{tabular}{|c|l|c|}
\hline \bf Transition & \bf Set of Words &\bf Regularity \\ \hline
 $<$null, ed$>$  &  succeed, seem,.. & Suffix  \\
 $<$null, ing$>$ & read, poison,..  & Suffix \\
 $<$ed,  ing$>$ &  viewed, documented,.. & Suffix \\
$<$able,null$>$ & sustainable, reasonable,.. & Suffix \\
$<$null, un$>$ & acceptably, accounted,.. & Prefix \\
\hline
\end{tabular}
\end{center}
\end{table}

\item Extract {\bf Transformation Rules}  from these transition sets - Transformation rules (see Table~\ref{font-table11}) capture word pairs which are morphologically similar (belong to same transition sets) apart from being syntactically similar according to initial word embeddings learned in SG.

{\bf Transformation Rules}:
For a word pair to be considered as a part of the transformation rules, both of its words should be frequent; a frequency threshold of 500 for small training sets and 1000 for large training set. This ensures that both the words of the pair are trained well. 

For every transition set, we find out which pair represents the maximum number of word pairs (the cosine similarity of their vector differences is above a threshold: 0.15 for prefix rules and 0.25 for suffix rules). If the count is greater than 10, we make it a transformation rule and remove it along with the other word pairs it represents from the transition set (because we want a limited number of morphological transformations representing similar morphological regularities). We then recursively follow the said technique for the transition set until we stop getting transformation rules from it. This approach results in multiple transformation rules from the same transition set. We see that it is needed because different forms of same transition exist (walk-walks, invention-inventions, object-objects - both verbs and nouns in this case).
\begin{table}[h]
\begin{center}
\caption{\label{font-table11}Examples of Transformation Rules }
\begin{tabular}{|c|c|}
\hline \bf Word Pair &\bf Regularity Type  \\ \hline
$<$side, beside$>$ &  Prefix  \\
$<$eighty, eight$>$ & Suffix \\
$<$transmitted, transmit$>$ & Suffix  \\
$<$cinematic, cinema$>$ & Suffix \\
$<$reminder, remind$>$ & Suffix \\
$<$after, afterward$>$ & Suffix \\
\hline
\end{tabular}
\end{center}
\end{table}
\end{itemize}
We observed that there were more impurities in words of smaller length. They were not removed at the time of extraction of transformation rules because these words are large in number. In this step, those rules in which either or both of the words have a length of less than or equal to three are eliminated. We also know that words which are morphological forms of each other have similar word embeddings and our aim is to create new sentences similar to the original sentence, so, we used a cosine similarity threshold of 0.1 for filtering the transformation rules.
\subsection{Morphological Sets}
\label{sect:pdf}
A morphological set (see Table~\ref{font-table7}) of a word is a collection of all morphological forms we detect. For words with frequency more than a threshold (100 for small and 1000 for large training sets were used in our experiments) and length greater than three, we attempt to construct morphological sets. We apply all the transformation rules on the word. For every resultant word with frequency more than or equal to 5 (hence has a trained word embedding in SG), we check if it is similar to the original word (cosine similarity greater than 0.15) and add it to the set.
\begin{table}[h]
\begin{center}
\caption{\label{font-table7}Examples of Morphological sets of some words }
\begin{tabular}{|l|l|}
\hline \bf Word & \bf Morphological Set  \\ \hline
comically &  comical, comic  \\
hanging & hang, overhanging, rehanging, hangings \\
woody & non-woody  \\
localized & unlocalized, localize, non-localized, local \\
cityhood & city \\
trawling & trawl \\
\hline
\end{tabular}
\end{center}
\end{table}
Now, we proceed to evaluating the morphological set. There may be many issues with the morphological set, homo-morphs being one. A homo-morph is a word which has more than one meaning. An example of a homo-morph is the word ``state" with its morph set \{stately, state-hood, stated, re-stated\}. We observe that using cosine similarity, we are able to eliminate such sets as they can be broken up into more than one subsets (\{stately, state-hood\}, \{stated,re-stated\}) whose words are not similar to each other. We select another random word from the set and assure that its cosine similarity with every word in the morph set is more than 0.15. This technique also removes morphologically unrelated words sharing a common prefix/suffix that may have crept in. For example, the word ``define" has \{definition, defined, fine\} as its morph set. We found that large morph sets usually contained more impurities. As we had only applied a single level of transformation for extracting these sets, we discard those sets whose size is greater than 6.
\subsection{Text Expansion}
\label{ssec:layout}
Now that we have an approximation of word similarities based on morphology, we would like to use this information to learn better word embeddings. We know that rare words' embeddings are not learned well due to lack of context. Based on words' morphological similarities we would want to overcome this shortfall by training embeddings on corpus augmented with artificially generated sentences. We want to generate new sentences using the morphological sets that we have generated. However, a brute force approach results in combinatorial explosion with a single sentence expanding to thousands of sentences. Therefore, we use a random function to choose which sentences to generate to prevent bias towards any specific morphological form. 

In this step, for each sentence in the training set, we attempt to create a maximum of one new sentence using a random function and morphological sets. For each word in the sentence, if that word has a morphological set, we choose with 50 percent probability to replace that word with a random word from its morph set. If the sentence thus generated is different from the original sentence, it becomes a part of our new training set upon which we train word embeddings in the next step.\\
In model SG + 2-Exp, we try to generate a maximum of two sentences from each sentence and a maximum of three new sentences in 3-Exp.
Shown below is a sample sentence of the original text.
\begin{quote}
``\textit{Maintenance is necessary for a software product to be successful
}''
\end{quote}
Now in expanded text (Exp).
\begin{quote}
``\textit{Maintenance is necessary for a software product to be successful\\
High-Maintenance is necessary for a software production to be successful}
''
\end{quote}
In expanded text (2-Exp).
\begin{quote}
``\textit{Maintenance is necessary for a software product to be successful\\
Maintenance is unnecessary for a software product to be successful\\
Maintenance is necessary for a software production to be unsuccessful
}''
\end{quote}
In expanded text (3-Exp).
\begin{quote}
``\textit{Maintenance is necessary for a software product to be successful\\
Self-Maintenance is unnecessary for a software product to be successful\\
Maintenance is unnecessary for a software production to be unsuccessful\\
High-Maintenance is necessary for a software production to be successful
}''
\end{quote}

We observe "necessary" and "successful" were replaced by their negation forms. This happened because their word vectors are similar which is due to the fact that they are often used interchangeably in natural language.
\subsection{Word Embeddings on Expanded Text (SG+Exp) }
We train word embeddings on the expanded text to train a new model. We will call this model SG + Exp (Skip-gram + Expansion). For exploring the effects of this method, we also train models SG + 2-Exp and SG + 3-Exp on the large dataset of English. For all the models tested, we have initialized vectors of unseen words with zero vectors.

\begin{table}[h]
\begin{center}
\caption{\label{font-table2} Small training samples (Spearman $\rho$). OOV represents Out of Vocabulary words encountered while testing.  }
\begin{tabular}{|c|l|c|l|c|l|}
\hline \bf Language & \bf Size &\bf SG &\bf OOV &\bf SG+EXP &\bf OOV \\ \hline
 Hindi  &  34M & 50.3  & 0 & 56.8 & 0\\
 Arabic & 34M  & 46.17 & 9 & 46.12 & 6\\
 Persian &  43M & 20.5 & 15 & 22.7 & 15\\
Spanish & 44M & 61 & 4 & 73.1 & 0\\
DE Gur & 44M & 38.8 & 54 & 47.5 & 32\\
DE RG & 44M & 14.7 & 1 & 28.2 &0\\
DE ZG & 44M & 22 & 74 & 27.8 &55\\
French & 39M & 48.2 & 3 & 61.4 &2\\
EN WS & 53M & 71.6 & 0 & 74.1 &0\\
EN RG & 53M & 64.5 & 0 & 71.3 &0\\
EN RW & 53M & 18.7 & 719 & 22 &255\\
\hline
\end{tabular}
\end{center}
\end{table}

\begin{table}[h]
\begin{center}
\caption{\label{font-table100} Large training samples (Spearman $\rho$). SO\cite{SO:15} and Size (SO) represents scores of ~\cite{SO:15} and size of datasets used respectively. These scores act as reference for our scores on small training samples.}
\begin{tabular}{|c|l|c|l|c|l|}
\hline \bf Language & \bf Size &\bf SG &\bf OOV &\bf  SO \cite{SO:15} &\bf Size (SO) \\ \hline
Hindi  & 0.75b & 61  & 0 & - & - \\
Arabic & 63M  & 48.3 & 4 & 43.1 & 0.45b \\
Persian &  59M & 22.1 & 15 & - & - \\
Spanish & 0.45b & 82.5& 0 & 47.3 & 0.56b \\
DE Gur & 0.5b & 57.1 & 23 & 64.1 & 1.2b \\
DE RG & 0.5b & 67.1 & 1 & - & - \\
DE ZG & 0.5b & 28.1& 38 & 21.5 & 1.2b \\
French & 0.2b & 64.6 & 1 & 67.3 & 1.5b \\
EN WS & 1b & 74.4 & 0 & 71.2 & 1.1b \\
EN RG & 1b & 77.9 & 0 & 75.1 & 1.1b \\
EN RW & 1b& 42.1 & 66 & 41.8 & 1.1b \\
\hline
\end{tabular}
\end{center}
\end{table}

\begin{table}[h]
\begin{center}
\caption{\label{font-table3}Trained on 1B tokens(English) - Comparison between different degrees of expansion (Spearman $\rho$)  }
\begin{tabular}{|l|c|c|c|}
\hline \bf System & \bf RW & \bf WS & \bf RG \\ \hline
SG & 42.1 & 74.4 & 77.9 \\
SG + Exp & 45.6 & 73.8 & 80.4\\
SG + 2-Exp & 45.9 & 72.9 & 80.3 \\
SG + 3-Exp & 45.3 & 71.3 & 78.3\\
\hline
\end{tabular}
\end{center}
\end{table}

Table \ref{rare} gives an insight into the effectiveness of the proposed method. We see that the words ``censorship" and its rarer morphological variant ``censorships" have very low similarity score and the similar words for ``censorships" are also not very informative. By training new embeddings on the expanded corpus we find that not only does the similarity score increase but the other moprhological variants also come closer to the word.

\begin{table}[h]
\begin{center}
\caption{\label{rare}Comparison between SG and SG + Exp regarding word pair $<$censorship,censorships$>$}
\begin{tabular}{|l|c|l|l|}
\hline \bf Model  & \bf Sim & \bf Closest to ``censorships"\\ \hline
SG  & 0.43 & POVs, normies \\
SG + Exp  & 0.87 & censorship, self-censorship \\
\hline
\end{tabular}
\end{center}
\end{table}

\section{Handling Rare and Unseen Words}
Since we assume rare words -- frequency lower than a threshold, have unreliable word embeddings, in this section, we explain how we build word embeddings of rare and unseen words at the time of evaluation using the transformation rules. Our approach relies on the regularities illustrated by Mikolov et. al. \cite{MSG2:13b} which were used successfully by Soricut and Och \cite{SO:15}.

\begin{table}[h]
\begin{center}

\caption{\label{font-table6}Tranined on 1B tokens(English) - Comparison between different systems after Morph step (Spearman $\rho$). Note that the scores on WS and RG are unchanged because the frequencies of all the words are above the threshold  }
\begin{tabular}{|l|c|c|c|}
\hline \bf System & \bf RW & \bf WS & \bf RG \\ \hline
SG + Morph & 45.1 & 74.4 & 77.9 \\
SG + Exp + Morph & 47.9 & 73.8 & 80.4\\
SG + 2-Exp + Morph & 47.6 & 72.9 & 80.3 \\
SG + 3-Exp + Morph & 46.8 & 71.3 & 78.3\\
\hline
\end{tabular}
\end{center}
\end{table}

In our experiments, we classify a word as rare if its frequency is less than 20. For building word embeddings for such words, we first find a reliable base word using Algorithm ~\ref{BaseAlgorithm}, from which rare word can be generated after successive application of transformation rule vectors.

\begin{algorithm}
\caption{Find Reliable Base Word}
\label{BaseAlgorithm}
\begin{algorithmic}[1]
\Procedure{Explore Level}{curWords}
\State $\textit{Words} \gets \text{Initialize with empty list}$
\For{each word $w$ \Pisymbol{psy}{206} $curWords$}
\For{each rule $r$ \Pisymbol{psy}{206} $Rules$}
\State $\textit{word}\gets\text{apply rule $r$ on $w$}$ 
\State $\textit{Words} \gets \text{append $word$}$
\EndFor
\EndFor
\State \Return{$Words$}
\EndProcedure
\Statex
\Procedure{Find Reliable Base}{rareWord}
\State $\textit{word}\gets\text{$rareWord$}$
\For{level $l$ \Pisymbol{psy}{206} $[1,3]$}
\State $\textit{Reliable} \gets \text{$l$*50}$
\State $\textit{words}\gets\text{$Explore Level(words)$}$
\State $\textit{FreqWord} \gets \text{MostFrequent($words$)}$
\If {$Freq[FreqWord]>\textit{Reliable}$} 
\State \Return $FreqWord$
\EndIf
\EndFor
\EndProcedure
\end{algorithmic}
\end{algorithm}

Our algorithm uses level order search, and returns us the most reliable base word (if any) along with the transformation rules required to generate our rare word from the base word. 

We increase our reliability threshold (Reliable) as the search level increases. This is done to because application of multiple transformation rules generally results in more errors. If even after searching for three levels we do not get our embedding, we use the embedding of the word itself if its present in the model (even though its probably poorly trained). In case the word is not present in the model, we try the above mentioned technique of level order search after capitalisation, followed by the lower case form of the word.

If we still do not have any embedding of the word, we recursively remove characters from the beginning and end of the word and generate all words with a frequency greater than 50 and a length greater than 3 (as the probability of error increases with small lengths) . We use the embedding of the word with the maximum length and use frequency in case of a tie. The same technique is applied on after capitalization and then the lowercase version of the word in case we do not get any embedding.

If we fail to get any embedding, we assign it a zero vector. This step is referred to as ``Morph". The system SG + Morph indicates that we have handled rare and unknown words using transformation rules when evaluating the model SG.
\begin{table}[h]
\begin{center}

\begin{tabular}{|l|c|c|c|c|}
\hline \bf System & \bf Size & \bf RW & \bf WS & \bf RG \\ \hline
LSM13~\cite{MAC:13} & 1B & 34.4 & 64.6 & 65.5 \\
 \hline
Glove~\cite{JRC:14} & 6B & 38.1 & 65.8 & 77.8 \\
Glove~\cite{JRC:14}& 42B & 47.8 & {\bf75.9} & {\bf82.9} \\
 \hline
SO15~\cite{SO:15} w/o M& 1.1B & 35.8 & 71.2 & 75.1\\
SO15~\cite{SO:15} w/ M& 1.1B & 41.8 & 71.2 & 75.1\\
\hline
LTM16~\cite{CHAR:16}& 0.3B & 47.1 & - & -\\
\hline
SG & 1B & 42.1 & 74.4 & 77.9 \\
SG + Morph & 1B & 45.1 & 74.4 & 77.9 \\
SG + Exp & 1B & 45.6 & 73.8 & 80.4\\
SG + Exp + Morph & 1B & {\bf47.9} & 73.8 & 80.4\\
\hline
\end{tabular}
\end{center}
\end{table}

\section{Result and Analysis}

The method seems to perform well on small datasets with an increase in accuracy on all datasets across all languages - English, German, French, Spanish, Persian and Hindi. Morphologically richer languages show more increase with respect to others. (see Table~\ref{font-table2}).

We see that the embeddings trained on a large dataset also tackles the data sparsity present in them (in the form of low frequency words). This is seen by the increase in the accuracies of RW and RG datasets which contain more low frequency words.
 
After analyzing these three datasets, we compute their rarity which is the average of the frequencies of the rarer word of the pair. The order of rarity of the datasets is RW(2253) $<$ RG(6505) $<$ WS(34152). The order helps in explaining the observation in terms of better word embeddings for low frequency words. For RW dataset, accuracy increases on expansion and the highest accuracy is found on SG + 2-Exp; for RG dataset, accuracy increases on expansion and the highest is found on SG + Exp. While for the WS dataset, accuracy decreases with any expansion. 

We observe that the accuracy of SG + Exp + Morph $>$ SG + 2-Exp + Morph (see Table~\ref{font-table6}) on RW dataset even though the accuracy of SG + 2-Exp $>$ SG + Exp (see Table~\ref{font-table6}). This happens because of decrease in analogical regularities as more artificial sentences are introduced. 

\section{Conclusion and Future Work}
\label{discussion}
Our method successfully exploits morphological regularities to produce high quality of word embeddings. Evaluations show that the method may be used to deal with the problem of data sparsity across different languages, the effect is particularly noticeable for morphologically rich languages. Our results are comparable to the ones presented earlier \cite{JRC:14} which was trained on a much larger dataset (42 billion tokens) compared to a much smaller training data (1 billion tokens) used in this method. 

 While synthesizing artificial sentences, a lot of sentences generated were grammatically incorrect. But since our initial word embeddings were poorly trained due to lack of data, these grammatically incorrect sentences do help in improving corresponding word embeddings since their main purpose is to provide more training samples, or artificial data for word2vec to train on. We will further explore how this artificially generated data might help us in other linguistic tasks, but for languages that are computationally poor, it helped in improving their word embeddings. We also evaluated our approach on languages that are resource rich, and as expected, word embeddings were affected negatively by our artificial sentences because initial word embeddings were already well trained. For languages that are resource rich, we will apply techniques like HMM or doc2vec on synthesized sentences to keep only those which are correct to a certain degree and study their impact on the word embeddings.  

As expected, the results for Arabic show the inherent complex nature of the language and can be accounted to the method being unable to reconcile infixes as opposed to suffixes and prefixes resulting in incorrect morphological sets. Future work will be directed towards generating  transformations which involve changes inside the word (apart from just changes to starting and endings)

\end{document}